# Deep Learning Enables Large-Scale Shape and Appearance Modeling in Total-Body DXA Imaging

Arianna Bunnell[1,2][0009-0000-6253-8402], Devon Cataldi[1][0000-0003-2958-6167], Yannik Glaser[2][0000-0001-7217-9749], Thomas K. Wolfgruber[1][0000-0001-8770-1800], Steven Heymsfield[3][0000-0003-1127-9425], Alan B. Zonderman[4][0000-0002-6523-4778], Thomas L. Kelly[5], Peter Sadowski[2][0000-0002-7354-5461] and John A. Shepherd[1][0000-0003-2280-2541]*

[1] University of Hawaiʻi Cancer Center, Honolulu HI, USA
`*jshepherd@cc.hawaii.edu`
[2] University of Hawaiʻi at Mānoa, Honolulu HI, USA
[3] Pennington Biomedical Research Center, Baton Rouge, LA, USA
[4] Laboratory of Epidemiology and Population Sciences, NIA/NIH/IRP, Baltimore MD, USA
[5] Hologic, Inc., Marlborough, MA, USA

**Abstract.** Total-body dual X-ray absorptiometry (TBDXA) imaging is a relatively low-cost whole-body imaging modality, widely used for body composition assessment. We develop and validate a deep learning method for automatic fiducial point placement on TBDXA scans using 1,683 manually-annotated TBDXA scans. The method achieves 99.5% percentage correct keypoints in an external testing dataset. To demonstrate the value for shape and appearance modeling (SAM), our method is used to place keypoints on 35,928 scans for five different TBDXA imaging modes, then associations with health markers are tested in two cohorts not used for SAM model generation using two-sample Kolmogorov-Smirnov tests. SAM feature distributions associated with health biomarkers are shown to corroborate existing evidence and generate new hypotheses on body composition and shape's relationship to various frailty, metabolic, inflammation, and cardiometabolic health markers. Evaluation scripts, model weights, automatic point file generation code, and triangulation files are available at https://github.com/hawaii-ai/dxa-pointplacement.

## 1 Introduction

Total-body dual X-ray absorptiometry (TBDXA) is clinically used for the diagnosis of osteoporosis, fracture risk assessment, and body composition analysis. TBDXA operates by emitting two distinct X-ray energy levels and measuring the attenuation at each pixel across the scanned image [1], allowing for pixel-level analysis of different tissue types. TBDXA represents a relatively low- cost, accessible whole-body imaging modality with low levels of radiation exposure for use in longitudinal studies [2].

Shape and Appearance Modeling (SAM) has been used extensively for the modeling of anatomy shape and texture specifically in medical imaging [3-5]. SAM uses principal components analysis (PCA) to mathematically represent the precise amount of variation found in shape and texture in a certain dataset with registered fiducial points or landmarks. The discovered principal components can be examined cross-sectionally (such as in investigating the shape of the human femur [6]), longitudinally (such as in determining bone maturity [7]), or as part of a data-generating process for downstream analysis (such as in classifying the gender of faces [8]). A challenge limiting the widespread adoption of SAM for studying the variance of shape is that it requires annotators, in medical imaging typically highly trained medical professionals, to place fiducial points on all images in the dataset. This makes preparation a costly and time-consuming endeavor for even small datasets. In this work, we develop an accurate method for automated point placement to facilitate the use of SAM on much larger TBDXA datasets.

Prior work has shown the importance of body shape for the prediction of health conditions and its influence on health marker values. Simple body shape measurements such as the body mass index (BMI), body shape index (ABSI), body roundness index (BRI), and trunk-to-leg volume, and waist-to-hip ratio have previously been shown to be predictive of diabetes [9-11], frailty [12], all-cause mortality [10, 12, 13], and cardiovascular events [14]. SAM from whole-body imaging modalities allows for more precise quantification of body shape than ABSI/BMI/BRI and does not require anthropometric measurements on subjects. [15] showed that features of body shape derived from SAM on 3-dimensional optical imaging (a modality which results in a registered 3-dimensional mesh of the body) were associated with cardiometabolic and metabolic health markers such as high-density lipoprotein (HDL) cholesterol, triglycerides, glucose, and insulin. This work builds on a previous study [16] which showed that TBDXA SAM features were predictive of mortality and associated with metabolic risk factors.

The main contributions of this paper are: (1) we define 105 bony and soft tissue fiducial points on TBDXA images; (2) we develop a deep learning method for automatic fiducial point placement on TBDXA imaging to facilitate shape and appearance modeling; (3) we show that differences in TBDXA SAM feature distributions correspond to differences in multiple metabolic and frailty-related health markers in an internal unseen testing dataset; and (4) we show that differences in TBDXA SAM feature distributions correspond to differences in multiple cardiometabolic, metabolic, and inflammation-related health markers in a new age demographic in an external cohort.

## 2  Methods and Dataset

### 2.1  Dataset

TBDXA images were collected from five different prospective National Institutes of Health studies: the Health, Aging, and Body Composition (HABC) study [17]; the Osteoporotic Fractures in Men (MrOS) study [18]; the Shape Up! Adults (SUA) study [15]; the Healthy Aging in Neighborhoods of Diversity across the Life Span (HANDLS) study [19]; Bone Mineral Density in Childhood (SUK) study [20] and the Multi-Ethnic Cohort (MEC) study [21]. SUK study data were only used for training the fiducial point placement model. HANDLS study data were exclusively used to examine association of SAM features with health markers. SUA and MEC study data were used for both fiducial point placement and SAM training. MROS and HABC study data were used for SAM training.

All scans included in this work were taken by Hologic scanners (QDR Series or Horizon A). TBDXA images were preprocessed using proprietary methods from Hologic Inc. which output tissue-specific data layers from raw projection and X-ray attenuation reference files [22]. This system applies energy discrimination and tissue decomposition algorithms based on the differential attenuation of low- and high-energy X-ray beams. From these files, the following mode-specific images were extracted using Hologic's internal material decomposition algorithm:

— Fat differential mode (`d_fat`): These images are generated by solving a set of simultaneous equations using known attenuation coefficients for fat at both energy levels. This mode isolates pixels attributable to adipose tissue. See **Figure 1A.**
— Bone mineral density mode (`bmd_irs`): These images provide pixel-by-pixel calibration data for deriving bone mineral content and density. See **Figure 1B.**
— Reference air mode (`r_air`): These images are used to correct for background attenuation and calibrate X-ray transmission in the absence of tissue. See **Figure 1C.**
— Lean differential mode (`d_lean`): These images isolate non-fat soft tissues, using attenuation coefficients characteristic of lean muscle and organs. See **Figure 1D.**
— Mass mode for bone tissue (`m_bone_irs`): These images represent bone-specific material density after calibration through the reference standard. See **Figure 1E.**

The Hologic decomposition algorithm applies calibration constants obtained from internal phantoms and system checks. These constants are used to translate differential attenuation readings into mass estimates of fat, lean, and bone tissue. Each pixel is classified through a least-squares fitting algorithm that assumes a three-compartment tissue model (bone, lean, fat), minimizing the residual error between predicted and observed attenuation. Additionally, scans are resized from their raw X-ray attenuation size (109 × 150 pixels) at 16-bit depth with a spatial resolution of 2mm×12.76mm per pixel. All images are upscaled with bicubic interpolation.

Availability of biomarkers varied. Subject height, weight, body mass index (BMI), age, sex, and race were available for all subjects in all studies. **Table 1** shows which biomarkers are available from which studies, means, and standard deviations.

**Table 1.** List of biomarkers. If the biomarker is available, the mean and (standard deviation) are shown. Variables with <90% completion are shown highlighted in gray. Inflam = Inflammation.

| | Biomarker (units) | HABC | | MrOS | HANDLS | |
|---|---|---|---|---|---|---|
| | | Males | Females | Males | Males | Females |
| Metabolic | Fasted insulin (uIU/mL) | 9.3 (14.3) | 9.3 (13.1) | 16.0 (9.2) | - | - |
| | Fasted blood glucose (mg/dL) | 105.8 (28.4) | 100.4 (28.1) | 100.4 (26.5) | - | - |
| | Hemoglobin A1c (%) | 6.06 (1.1) | 5.98 (1.0) | - | 6.10 (1.4) | 6.07 (1.2) |
| | Blood glucose (non-fasted) (mg/dL) | 106.2 (29.2) | 100.6 (28.6) | - | 109.3 (46.1) | 103.7 (36.1) |
| Frailty | Gamma glutamyl transferase (U/L) | - | - | - | 57.0 (116.9) | 40.1 (66.1) |
| | Magnesium (mg/dL) | - | - | - | 1.99 (0.2) | 1.98 (0.2) |
| | Grip strength (kg) | 34.9 (8.1) | 21.4 (5.7) | 38.4 (8.8) | - | - |
| | Long (400 m) walk speed (m/sec) | 1.30 (1.06) | 1.21 (1.10) | 1.02 (0.20) | - | - |
| | Short (3/4/6 m) walk speed (m/sec) | 1.19 (0.24) | 1.08 (0.23) | 1.16 (0.24) | - | - |
| Cardiometa- | HDL cholesterol (mg/dL) | - | - | - | 52.7 (19.4) | 59.5 (18.5) |
| | Cholesterol/HDL ratio | - | - | - | 3.71 (1.4) | 3.51 (1.2) |
| | Calcium (mg/dL) | - | - | - | 9.41 (0.4) | 9.44 (0.5) |
| | Creatinine (mg/dL) | - | - | - | 1.09 (0.8) | 0.86 (0.5) |
| Inflam. | C-reactive protein (mg/L) | - | - | - | 4.59 (9.1) | 6.9 (10.9) |
| | Platelet count ($10^9$/L) | - | - | - | 225.2 (63.6) | 258.8 (72.5) |

## 2.2   Automated Fiducial Point Placement Method

We defined 105 fiducial points on the skin edge and bone landmarks, adapted from prior work [16]. Distal limbs were excluded from point definitions due to a high likelihood of incidental positional differences. See **Figure 1** for a visualization of all 105 points on both a female and male for all TBDXA modes. Predefined, sex-specific triangulated meshes used for alignment and warping and are included in the released code.

A subset of 1,683 scans from the MEC, SUA, and SUK studies were previously manually annotated for the 105 fiducial points by trained research assistants by the investigators. Data were randomly split into training (80%; n = 1,348) and validation (20%; n = 335) sets by subject. No subjects or scans are represented in more than one split. A random selection of 500 HABC scans were annotated by a single expert (D. C.) to serve as an external testing set for the automated fiducial point placement method. Fiducial points were placed on the `r_air` images (see **Figure 1C**) as this channel provides an acceptable balance of soft tissue and bone visibility.

Manual fiducial point placement is a labor-intensive process requiring specialized training and taking an average of 2 minutes/scan to complete. To facilitate automated fiducial point placement on the large TBDXA dataset included in this work (n = 35,928 scans), we adapt methods used for 3D human pose estimation. We finetune a DeepPose [23] model with a ResNet-101 [24] backbone pretrained with residual log-likelihood estimation [25] on MS-COCO [26]. Models were evaluated using the percentage of correct keypoints (PCK), end point error (EPE), and normalized mean error (NME). PCK measures the average accuracy of all placed fiducial points, where a point placed within a normalized distance threshold (0.1) from the true point is considered correctly predicted. EPE measures the average Euclidean distance (in pixels) between the predicted and ground-truth fiducial points. NME measures the Euclidean distance (in pixels), normalized by the size of the bounding box containing all fiducial points. EPE, PCK, and NME on the validation set are compared to point placement previously computed from a random forest-based method [16] based on the AM_TOOLS pipeline.

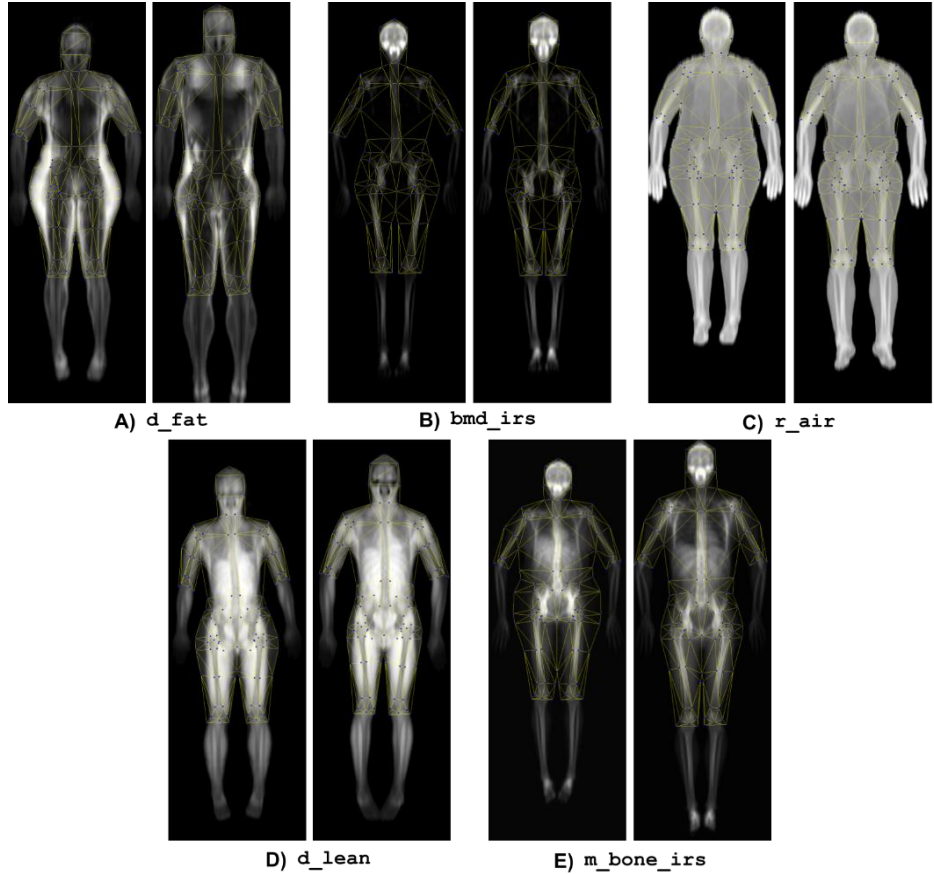

**Fig. 1.** Illustration of 105 placed fiducial points for statistical appearance modeling (blue) and predefined sex-specific triangulations (yellow). Each sub-figure shows a single extracted TBDXA imaging mode. Each sub-figure shows a female (left) and male (right) scan.

### 2.3   Statistical Appearance Modeling

Data were split via study and to preserve consistency [27] into training and testing sets for SAM. HANDLS was used to test association of discovered appearance principal components with biomarkers of interest in a completely independent sample. **Table 2** and **Table 3** show training and testing set characteristics by study. The HANDLS dataset had a median participant age of 55.1 years with 47.6 years as the 25th percentile and 61.6 years as the 75th percentile. The HABC dataset had a median participant age of 75.0 years with 73.0 years as the 25th percentile and 77.0 years as the 75th percentile.

Table 2. Characteristics of the SAM training study sample.

|  | Overall | HABC | MEC | MrOS | SUA |
|---|---|---|---|---|---|
| **Participants, N** | 9,597 | 2,403 | 833 | 5,939 | 422 |
| Female participants, N | 1,881 | 1,255 | 395 | 0 | 231 |
| Male participants, N | 7,716 | 1,148 | 438 | 5,939 | 191 |
| Mean age, years (SD) | 74.5 (6.2) | 75.3 (3.0) | 69.7 (2.8) | 75.1 (5.8) | 45.2 (16.1) |
| Mean height, cm (SD) | 169.3 (9.3) | 165.9 (9.5) | 165.3 (9.7) | 173.6 (6.9) | 168.1 (9.5) |
| Mean weight, kg (SD) | 78.2 (14.9) | 74.8 (15.1) | 76.9 (16.4) | 82.3 (13.2) | 76.6 (21.1) |
| **TBDXA scans, N** | 26,462 | 13,617 | 833 | 11,587 | 425 |
| Mean scans/person, N (SD) | 2.8 (2.1) | 5.7 (2.2) | 1.0 (0.0) | 2.0 (0.7) | 1.0 (0.1) |
| Female participant scans, N | 7,810 | 7,181 | 395 | 0 | 234 |
| Male participant scans, N | 18,652 | 6,436 | 438 | 11,587 | 191 |
| Mean BMI, kg/cm$^2$ (SD) | 27.2 (4.5) | 27.2 (4.8) | 28.1 (5.0) | 27.3 (3.8) | 27.0 (6.6) |
| Underweight scans, N | 290 | 235 | 5 | 23 | 27 |
| Normal scans, N | 8,172 | 4,460 | 239 | 3,322 | 151 |
| Overweight scans, N | 11,841 | 5,505 | 342 | 5,855 | 139 |
| Obese scans, N | 6,159 | 3,417 | 247 | 2,387 | 108 |

Table 3. Characteristics of the SAM testing study sample.

|  | Overall | HABC | HANDLS |
|---|---|---|---|
| **Participants, N** | 3,255 | 639 | 2,616 |
| Female participants, N | 1,819 | 318 | 1,499 |
| Male participants, N | 1,436 | 321 | 1,113 |
| Mean age, years (SD) | 62.6 (12.6) | 75.2 (3.0) | 54.9 (9.6) |
| Mean height, cm (SD) | 167.8 (9.7) | 166.3 (10.0) | 168.6 (9.3) |
| Mean weight, kg (SD) | 82.6 (20.6) | 75.2 (14.7) | 87.1 (22.3) |
| **TBDXA scans, N** | 9,466 | 3,602 | 5,861 |
| Mean scans/person, N (SD) | 2.91 (1.8) | 5.64 (2.2) | 2.24 (0.9) |
| Female participant scans, N | 5,241 | 1,809 | 3,432 |
| Male participant scans, N | 4,222 | 1,793 | 2,429 |
| Mean BMI, kg/cm$^2$ (SD) | 29.3 (6.9) | 27.1 (4.7) | 30.7 (7.7) |
| Underweight scans, N | 205 | 55 | 150 |
| Normal scans, N | 2,396 | 1,162 | 1,234 |
| Overweight scans, N | 3,187 | 1,575 | 1,612 |
| Obese scans, N | 3,675 | 810 | 2,865 |

Deep Learning Enables Large-Scale Shape and Appearance Modeling in TBDXA Imaging       7SAM was undertaken using the method developed in [28] and initially applied to TBDXA images in [16]. SAM and display functions were done in C++ using the UoMApM software [29]. Models of appearance were built for the `d_lean, bmd_irs, m_bone_irs, d_fat` TBDXA imaging modes. Models of shape are displayed on the `r_air` mode only. All models were constructed to explain 95% of the variation in the shape, or combined shape and appearance, in the training dataset.

In brief, the SAM method involves building a registered shape model from placed fiducial points, followed by building a texture model on the triangulated, mean discovered body shape. The appearance model consists of PCA jointly applied to both discovered shape and appearance vectors, forming a single representation of body shape and the tissue mode being imaged. The shape model is constructed by: (1) translation of fiducial points to a constant center of gravity; and (2) application of PCA to aligned fiducial points. The texture model is constructed by: (1) warping of images via affine transformation on the predefined triangulated mesh; (2) application of PCA to the grayscale values inside the body. All models were sex-specific to avoid the constructed principal component space becoming saturated with sex-specific variation.

Association of the discovered SAM features with health markers was undertaken using two-sample Kolmogorov-Smirnov (KS) tests [30]. For each feature in each SAM space (shape, `d_lean, bmd_irs, m_bone_irs,` and `d_fat`) all images in the test dataset falling below the 10th and above the 90th percentile of the principal component in the training dataset were identified. For each health marker, a two-sample KS test was performed with the null hypothesis that the two samples (low and high principal component) were drawn from the same underlying distribution of the selected health marker. Our hypothesis is that the tails of the principal component distributions contain information about general health and aging, as measured by available health markers (see Table 1). The 90th and 10th percentiles were defined on the HABC (females), or HABC/MROS (males) training sets and KS tests were performed in the HABC testing set. Additionally, Spearman's correlation coefficients were calculated in the training set between the shape SAM features and health markers. We use previously published interpretation ranges for Spearman's $\rho$ [31]. All tests on the HABC dataset were performed at Bonferroni-adjusted significance level of $\alpha = 4.1 \times 10^{-6}$.

To test transferability of the discovered shape and appearance features, we also investigated their distributional association with health markers in the HANDLS dataset. The HANDLS dataset is a much younger population with different health markers available. The shape and appearance features were not derived on the HANDLS dataset for any mode or sex. Like the procedure in the internal HABC testing set, association of the principal components with health markers was undertaken with two-sample KS tests, where the 10th and 90th percentile (low and high feature distributions) are defined on the HANDLS dataset. As a final test of transferability, the health markers Hemoglobin A1c and Glucose (non-fasted) were tested on both the HANDLS and HABC test datasets. All tests on the HANDLS dataset were performed at Bonferroni-adjusted significance level of $\alpha = 1.3 \times 10^{-6}$. Representative images were constructed for all health markers by taking the mean of the principal component at the 10th and 90th percentile if significant, or over the entire dataset if not. If less than 100 scans were present in either of the datasets, KS tests were not performed.

## 3   Results

### 3.1   Fiducial Point Placement

Model hyperparameters were systematically searched over 10 trials using the TPESampler in Optuna [32]. The final model was chosen based on PCK on the validation set. All models were trained using the Adam [33] optimizer with the following augmentations: random bounding box (defined by keypoint boundaries) rotation and scaling, random flipping, and random half-body masking. The hyperparameter search space contained the learning rate ($1\times10^{-7}$, $1\times10^{-2}$) and the learning rate scheduler. The selected learning rate was $9\times10^{-6}$ and the learning rate was reduced by half on plateau during training. Fiducial point placement results on the set of 500 scans from the external HABC dataset were: 99.5% PCK, 8.49 EPE, and 0.007 NME. On the validation set results were: 99.9% PCK, 4.48 EPE, and 0.004 NME as compared to the method from [16] which performed at 97.1% PCK, 40.9 EPE, and 0.037 NME. Predictions were generated from the trained model for the MROS, HANDLS, and the remaining HABC scans, resulting in a total of 34,170 sets of fiducial points being placed by the model.

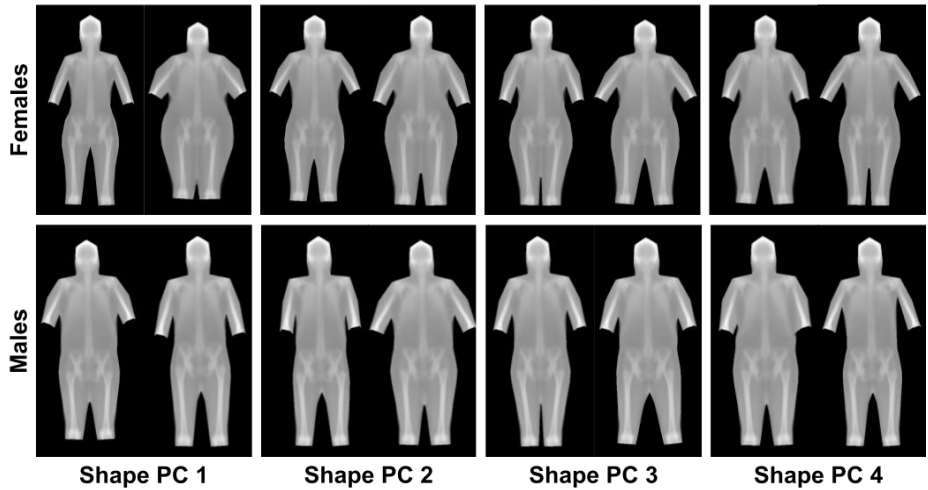

**Fig. 2.** Visualization of the first 6 principal components (PC) at +/- 2.5 standard deviations from the shape and appearance models (SAM) for males and females on the `r_air` TBDXA mode.

### 3.2   Shape and Appearance Modeling

We found that 29 shape modes explained 95% of the shape variance for males and 28 shape modes explained 95% of the shape variance for females. For females, the number of principal components required to explain 95% of appearance (combined texture and shape) variance were: 66 for `d_fat`, 89 for `bmd_irs`, 179 for `d_lean`, and 125 for `m_bone_irs`. For males, the number of principal components required to explain 95% of appearance (combined texture and shape) variance were: 59 for `d_fat`, 71 for

Deep Learning Enables Large-Scale Shape and Appearance Modeling in TBDXA Imaging  9

`bmd_irs`, 209 for `d_lean`, and 115 for `m_bone_irs`. The first 4 shape modes at +/-2.5 standard deviations from the mean, visualized on `r_air` images, for each sex are shown in **Figure 2.** Note the principal components contain information about both body shape and posing differences. For example, arm positioning varies in principal component 2 and leg positioning varies in principal component 3 for males. Body shape differences are seen very clearly in principal components 1, 2, and 3 for females.

Spearman's correlations between the top five shape SAM features and the metabolic/frailty health markers available in the HABC training dataset are shown in **Table 4.** Only correlation coefficients significant at the α = 0.01 level are shown in the table. One shape SAM feature showed moderate association with fasted insulin in both males and females. The remaining health markers – fasted blood glucose, hemoglobin A1c, grip strength, long walk speed, and short walk speed – showed weak association with SAM features in both sexes. Shape features related to body size (PC 1 females/PC 4 males) were associated with fasted insulin and hemoglobin A1c.

**Table 4.** Spearman's correlation coefficients between the first 10 discovered shape principal components and health markers in the HABC SAM training set. HbA1c = Hemoglobin A1c.

|  | Shape SAM Feature | Fasted insulin | Fasted blood glucose | HbA1c | Grip strength | Long walk speed | Short walk speed |
|---|---|---|---|---|---|---|---|
| Females | PC 1 | 0.396 | 0.256 | 0.152 |  | -0.143 | -0.145 |
|  | PC 2 |  |  |  |  | -0.206 | -0.183 |
|  | PC 3 |  |  |  | -0.110 |  |  |
|  | PC 4 | -0.135 |  |  | -0.153 |  |  |
|  | PC 5 | 0.131 |  |  |  |  |  |
| Males | PC 1 | -0.199 | -0.158 |  |  |  |  |
|  | PC 2 | -0.142 | 0.105 |  | -0.167 |  | -0.110 |
|  | PC 3 |  |  |  |  |  |  |
|  | PC 4 | -0.258 |  | -0.123 |  | 0.114 |  |
|  | PC 5 |  |  |  |  |  |  |

KS tests for health marker distributional differences between the high and low principal component groups had varying effects depending on the health marker and sex. **Figure 3** and **Figure 4** show the result of KS tests for the HABC and HANDLS tests sets on males and females, respectively. Each bar shows the discovered features for a TBDXA mode's shape and appearance model, and each group of bars represents a single marker. Health markers are grouped by indication and study (see **Table 1**).

In the HABC study, frailty markers were generally not distributed differently between the low and high tails. A notable exception to this is short walking speed in females, which was found to have differing distributions at the extremes for principal components explaining between 20-40% of total variance in the `bmd_irs`, `d_lean`, and `d_fat` modes. Differences in distribution of the metabolic marker fasted insulin were found in both females and males, across almost all TBDXA imaging modes. Fasted glucose was found to be differently distributed for the extremes of `d_fat` for females and males but had generally more differences found in females.

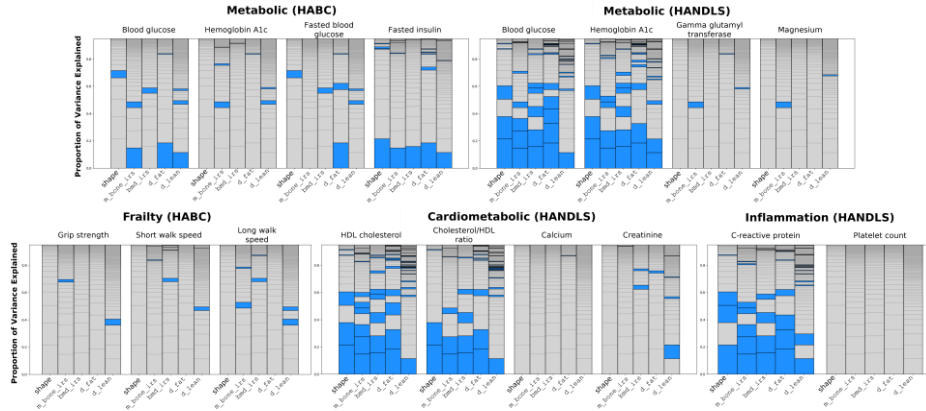

**Fig. 3.** Grouped bar charts showing the proportion of variance explained by each principal component for each SAM model (shape/appearance) for all TBDXA imaging modes for males. Figure is best viewed in color. Blue highlighting indicates significant differences.

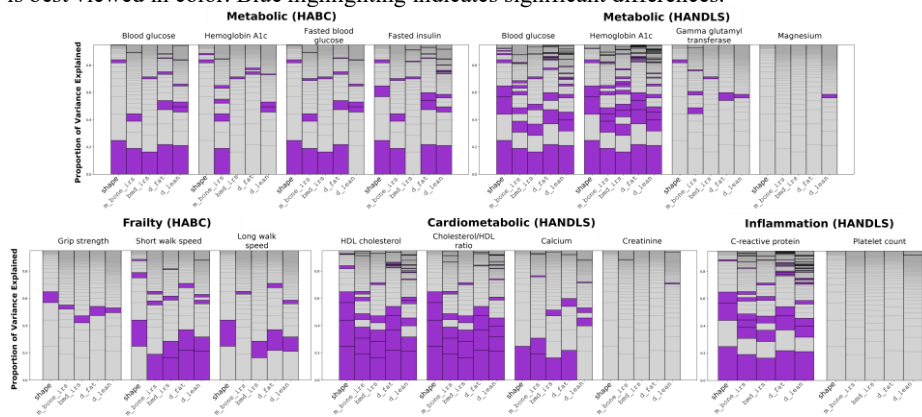

**Fig. 4.** Grouped bar charts showing the proportion of variance explained by each principal component for each SAM model (shape/appearance) for all TBDXA imaging modes for females. Figure is best viewed in color. Purple highlighting indicates significant differences.

In the younger external HANDLS cohort, cardiometabolic markers HDL cholesterol and Cholesterol/HDL ratio were found to be differently distributed by the KS tests for the tails of the SAM features explaining at least 50% of total variance in appearance in the shape, `bmd_irs`, `d_fat`, and `m_bone_irs` modes for both males and females. `d_lean` also showed differences in females only. SAM features did not reveal differences in Calcium or Creatinine in males and showed distributional differences with only Calcium (20-30% of variance) in females. Gamma glutamyl transferase, Magnesium, and Platelet count had few differences in distribution. The tails of SAM features across all TBDXA modes and for males and females were associated with significantly different distributions in the inflammation marker C-reactive protein.

The associations between non-fasted blood Glucose and Hemoglobin A1c were notably different between the older HABC set and the younger HANDLS set. In females, the first SAM features were associated with differences at the tails for non-fasted blood Glucose across both sets, but with fewer features showing association in the HABC set. In males, fasted blood Glucose showed a much higher proportion of variance associated with distributional differences (40-60%) in HANDLS than in HABC.

**Figure 5** shows representative images for short walk speed, non-fasted blood Glucose, calcium, C-reactive protein, fasted blood Glucose, and hemoglobin A1c. Representative images are built by taking the mean of the feature at the $10^{th}$ and $90^{th}$ percentile if significant, or over the entire dataset if not, and transforming back to image space.

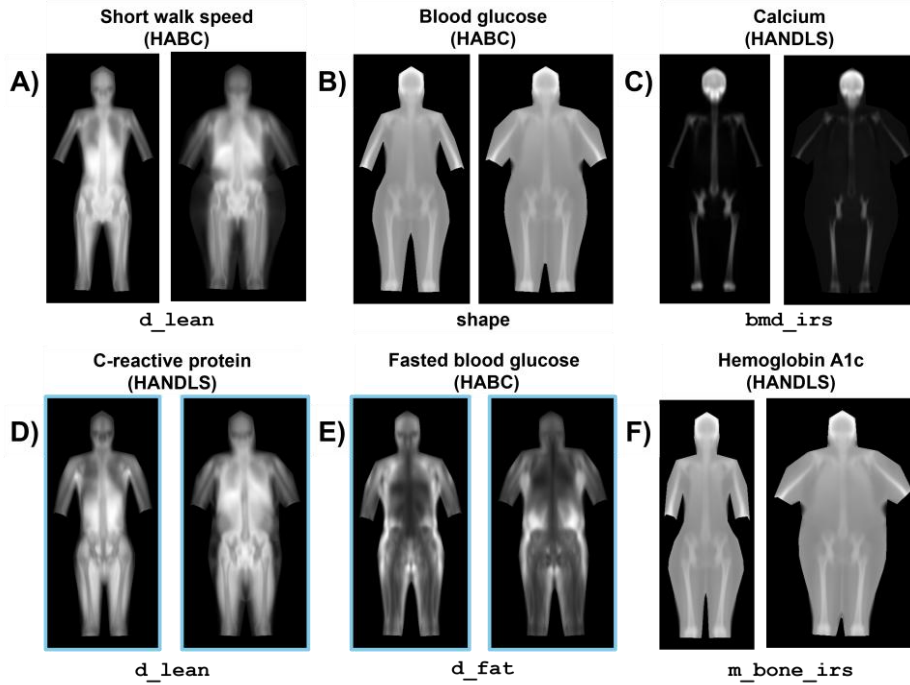

**Fig. 5.** Reconstructed images from the $10^{th}$ (left) and $90^{th}$ (right) percentiles for significant SAM features. Male reconstructions are shown with a blue outline. Other reconstructions are females.

## 4    Discussion

Our automated placement of fiducial markers is highly accurate and enables us to systematically annotate TBDXA images. We have demonstrated this by extracting SAM features from a large dataset of TBDXA data and performing statistical analysis to identify and visualize body composition features that are associated with health markers. This analysis enables hypothesis building and data-driven discovery, and in this section, we explore some of the hypotheses that resulted from this exploratory analysis.

**Figure 5A** shows representative reconstructed images at the $10^{th}$ (left) and $90^{th}$ (right) percentile of the `d_lean` SAM features statistically significantly associated with short walk speed in females. Slow walking speed is a frailty marker [34] which been shown to be associated with all-cause mortality [35, 36], cardiovascular death [36], and incident disability [37] in older adults. The $10^{th}$ percentile image (slow walk speed) shows a small body habitus with smaller limb circumference, as well as lower lean mass in the limbs. Arm circumference has been explored as a frailty marker in men [38], but has shown mixed results for functional ability in studies including females [39, 40]. Discovered SAM features relating to walk speed from the HABC cohort of older female adults may suggest additional investigation is warranted.

Body size appears to be strongly associated with distributional differences in metabolic health markers like blood glucose (**Figure 5B**), fasted blood glucose (**Figure 5E**), and hemoglobin A1c (**Figure 5F**), in both sexes, across both testing datasets. For these markers, the $90^{th}$ percentile PC image consistently shows larger body habitus than the $10^{th}$ percentile PC image. Additionally, for these markers the $90^{th}$ percentile image consistently appears more "apple-shaped" than the $10^{th}$ percentile image, with increased waist circumference. Increased waist circumference is one of the metabolic abnormalities needed for diagnosis of metabolic syndrome [41]. High calcium levels have also been shown to be associated with high levels of blood glucose and total cholesterol (see **Figure 5C**). **Figure 5E** also appears to show increased fat mass accumulated around the abdomen of the scan. Waist circumference and visceral adipose tissue thickness have been shown to be strongly associated with indicators of metabolic syndrome [42]. Additional examination of exact fat distributions through SAM feature analysis may reveal more predictive relationships. [43] investigates the relationship between visceral adipose tissue distribution and metabolic syndrome through both abdominal MRI and anthropometry, representing higher examination cost than TBDXA and SAM analysis.

The marker C-reactive protein is produced in the liver in response to immune cells releasing proinflammatory cytokines. Plaque build-up in the arteries, atherosclerosis, is a chronic inflammatory disease which is a major cause of cardiovascular disease [44]. Increased C-reactive protein levels have been shown to be strongly predictive of adverse cardiovascular events and coronary heart disease [45, 46]. **Figure 5D** shows larger body habitus, increased waist circumference, and decreased proportion of lean tissue, particularly in the trunk and the arms. Preliminary work in the NHANES cohort has demonstrated a link between arm fat mass and cardiovascular disease risk [47].

## 5    Conclusion

Examination of SAM features in TBDXA imaging may reveal relationships between precise measures of body composition and shape and health markers. In this work, we present and publicly release a deep learning solution for placing fiducial points to allow SAM on large datasets of TBDXA imaging, perform SAM on a TBDXA dataset of over 30,000 images, and show that discovered SAM features may be used to support and generate hypotheses about body composition and shape. The limitations of this work are a lack of consistent health markers across studies, limiting cross-study analyses; a large difference in the age composition of the internal and external testing sets, limiting direct evidence of transferability to new cohorts of older adults; and lack of longitudinal analysis examining the mutual changes in SAM features and health markers. Future work may use a more comprehensive preprocessing method for pose alignment, potentially isolating the SAM features to discovery of biological or body shape differences.

**Acknowledgements.** The Osteoporotic Fractures in Men Study is supported by National Institutes of Health (NIH) funding. The following institutes provide support: the National Institute of Arthritis and Musculoskeletal and Skin Diseases (NIAMS), the National Institute on Aging (NIA), the National Center for Research Resources (NCRR), and NIH Roadmap for Medical Research under the following grant numbers: U01 AR45580, U01 AR45614, U01 AR45632,U01 AR45647, U01 AR45654, U01 AR45583, U01 AG18197, U01-AG027810, and UL1 RR024140. The Shape Up! Kids study is supported by NIH NORC Center Grants P30DK072476, Pennington/Louisiana, P30DK040561, Harvard, and R01DK111698, Shape UP! Kids. The Shape Up! Adults study is supported by NIH R01 DK109008. The Healthy Aging in Neighborhoods of Diversity across the Life Span study is supported by the Intramural Research Program, NIA, NIH, grant Z01-AG000513. The Health, Aging, and Body Composition study is supported by NIA Contracts N01-AG-6-2101; N01-AG-6-2103; N01-AG-6-2106; NIA grant R01-AG028050, and NINR grant R01-NR012459. This research was funded in part by the Intramural Research Program of the NIH, NIA.

**Disclosure of Interests.** JAS is the recipient of an IIT grant from Hologic Inc. which partially supported this work. The remaining authors have no competing interests to declare.

# References

1. Heymsfield, S., Human body composition. Vol. 918. 2005: Human kinetics.
2. Blake, G.M., M. Naeem, and M. Boutros, Comparison of effective dose to children and adults from dual X-ray absorptiometry examinations. Bone. **38**(6): p. 935-942 (2006).
3. Cootes, T.F. and C.J. Taylor. Statistical models of appearance for medical image analysis and computer vision. Medical Imaging 2001: Image Processing. 2001. SPIE.
4. Cootes, T.F., et al., Use of active shape models for locating structures in medical images. Image and vision computing. **12**(6): p. 355-365 (1994).
5. Patenaude, B., et al., A Bayesian model of shape and appearance for subcortical brain segmentation. Neuroimage. **56**(3): p. 907-922 (2011).
6. Bryan, R., et al., Statistical modelling of the whole human femur incorporating geometric and material properties. Medical engineering & physics. **32**(1): p. 57-65 (2010).
7. Thodberg, H.H., et al., The BoneXpert method for automated determination of skeletal maturity. IEEE transactions on medical imaging. **28**(1): p. 52-66 (2008).
8. Makinen, E. and R. Raisamo, Evaluation of gender classification methods with automatically detected and aligned faces. IEEE transactions on pattern analysis and machine intelligence. **30**(3): p. 541-547 (2008).
9. Fujita, M., et al., Predictive power of a body shape index for development of diabetes, hypertension, and dyslipidemia in Japanese adults: a retrospective cohort study. PLoS One. **10**(6): p. e0128972 (2015).

10. Wilson, J.P., et al., Ratio of trunk to leg volume as a new body shape metric for diabetes and mortality. PLoS One. **8**(7): p. e68716 (2013).
11. Bawadi, H., et al., Body shape index is a stronger predictor of diabetes. nutrients. **11**(5): p. 1018 (2019).
12. Zhang, J. and H. Zhang, The association of body roundness index and body mass index with frailty and all-cause mortality: a study from the population aged 40 and above in the United States. Lipids in Health and Disease. **24**(1): p. 30 (2025).
13. Krakauer, N.Y. and J.C. Krakauer, A new body shape index predicts mortality hazard independently of body mass index. PloS one. **7**(7): p. e39504 (2012).
14. Lam, B.C.C., et al., Comparison of body mass index (BMI), body adiposity index (BAI), waist circumference (WC), waist-to-hip ratio (WHR) and waist-to-height ratio (WHtR) as predictors of cardiovascular disease risk factors in an adult population in Singapore. PloS one. **10**(4): p. e0122985 (2015).
15. Ng, B.K., et al., Detailed 3-dimensional body shape features predict body composition, blood metabolites, and functional strength: the Shape Up! studies. The American journal of clinical nutrition. **110**(6): p. 1316-1326 (2019).
16. Shepherd, J.A., et al., Modeling the shape and composition of the human body using dual energy X-ray absorptiometry images. PLOS ONE. **12**(4): p. e0175857 (2017).
17. Newman, A.B., et al., Strength and muscle quality in a well-functioning cohort of older adults: the Health, Aging and Body Composition Study. Journal of the American Geriatrics Society. **51**(3): p. 323-330 (2003).
18. Orwoll, E., et al., Design and baseline characteristics of the osteoporotic fractures in men (MrOS) study—a large observational study of the determinants of fracture in older men. Contemporary clinical trials. **26**(5): p. 569-585 (2005).
19. Evans, M.K., et al., Healthy aging in neighborhoods of diversity across the life span (HANDLS): overcoming barriers to implementing a longitudinal, epidemiologic, urban study of health, race, and socioeconomic status. Ethnicity & disease. **20**(3): p. 267 (2010).
20. Kalkwarf, H.J., et al., The bone mineral density in childhood study: bone mineral content and density according to age, sex, and race. The journal of clinical endocrinology & metabolism. **92**(6): p. 2087-2099 (2007).
21. Kolonel, L.N., et al., A multiethnic cohort in Hawaii and Los Angeles: baseline characteristics. American journal of epidemiology. **151**(4): p. 346-357 (2000).
22. Prof. Sambrook, P., The Evaluation of Osteoporosis. GM Blake, HW Wahner, and I. Fogelman, Editors. Martin Dunitz, London, UK, 1999. 1999, John Wiley and Sons and The American Society for Bone and Mineral Research ….
23. Toshev, A. and C. Szegedy. Deeppose: Human pose estimation via deep neural networks. Proceedings of the IEEE conference on computer vision and pattern recognition. 2014.
24. He, K., et al. Deep residual learning for image recognition. Proceedings of the IEEE conference on computer vision and pattern recognition. 2016.
25. Li, J., et al. Human pose regression with residual log-likelihood estimation. Proceedings of the IEEE/CVF international conference on computer vision. 2021.
26. Lin, T.-Y., et al., Microsoft COCO: Common Objects in Context, in Computer Vision – ECCV 2014. 2014, Springer International Publishing. p. 740-755.
27. Glaser, Y., et al., Deep learning predicts all-cause mortality from longitudinal total-body DXA imaging. Communications Medicine. **2**(1): p. 102 (2022).

28. Cootes, T.F., G.J. Edwards, and C.J. Taylor, Active appearance models. IEEE Transactions on pattern analysis and machine intelligence. **23**(6): p. 681-685 (2001).
29. Cootes, T.F., University of Manchester Appearance Model Library and Tools. 2024.
30. Massey Jr, F.J., The Kolmogorov-Smirnov test for goodness of fit. Journal of the American statistical Association. **46**(253): p. 68-78 (1951).
31. Schober, P., C. Boer, and L.A. Schwarte, Correlation coefficients: appropriate use and interpretation. Anesthesia & analgesia. **126**(5): p. 1763-1768 (2018).
32. Akiba, T., et al. Optuna: A Next-generation Hyperparameter Optimization Framework. International Conference on Knowledge Discovery and Data Mining. ACM.
33. Kingma, D.P. and J. Ba, Adam: A method for stochastic optimization. arXiv preprint arXiv:1412.6980, (2014).
34. Syddall, H.E., et al., Self-reported walking speed: a useful marker of physical performance among community-dwelling older people? Journal of the American Medical Directors Association. **16**(4): p. 323-328 (2015).
35. Liu, B., et al., Usual walking speed and all-cause mortality risk in older people: A systematic review and meta-analysis. Gait & posture. **44**: p. 172-177 (2016).
36. Dumurgier, J., et al., Slow walking speed and cardiovascular death in well functioning older adults: prospective cohort study. Bmj. **339** (2009).
37. Woo, J., Walking speed: a summary indicator of frailty? Journal of the American Medical Directors Association. **16**(8): p. 635-637 (2015).
38. Kim, S.Y., et al., Mid-upper arm circumference as a screening tool for identifying physical frailty in community-dwelling older adults: The Korean Frailty and Aging Cohort Study. Geriatrics & Gerontology International. **24**(12): p. 1292-1299 (2024).
39. Tsai, H.-J. and F.-K. Chang, Associations between body mass index, mid-arm circumference, calf circumference, and functional ability over time in an elderly Taiwanese population. PloS one. **12**(4): p. e0175062 (2017).
40. Xu, L., et al., Association between body composition and frailty in elder inpatients. Clinical interventions in aging: p. 313-320 (2020).
41. Swarup, S., et al., Metabolic syndrome, in StatPearls [internet]. 2024, StatPearls Publishing.
42. Kawada, T., T. Andou, and M. Fukumitsu, Waist circumference, visceral abdominal fat thickness and three components of metabolic syndrome. Diabetes & Metabolic Syndrome: Clinical Research & Reviews. **10**(1): p. 4-6 (2016).
43. Demerath, E.W., et al., Visceral adiposity and its anatomical distribution as predictors of the metabolic syndrome and cardiometabolic risk factor levels. The American journal of clinical nutrition. **88**(5): p. 1263-1271 (2008).
44. Björkegren, J.L. and A.J. Lusis, Atherosclerosis: recent developments. Cell. **185**(10): p. 1630-1645 (2022).
45. Shrivastava, A.K., et al., C-reactive protein, inflammation and coronary heart disease. The Egyptian Heart Journal. **67**(2): p. 89-97 (2015).
46. Ridker, P.M. and D.A. Morrow, C-reactive protein, inflammation, and coronary risk. Cardiology clinics. **21**(3): p. 315-325 (2003).
47. Liu, Z., et al., Relationship between body regional fat and cardiovascular disease in middle-aged and young adults: Results from NHANES 2011–2018 and Mendelian randomization meta-analysis. medRxiv: p. 2024.03. 19.24304562 (2024).